\documentclass[sigconf]{acmart}

\usepackage{graphicx} 
\usepackage{subfigure}

\usepackage{amsmath, amssymb}
\usepackage{makecell}
\usepackage{multirow}

\AtBeginDocument{%
  \providecommand\BibTeX{{%
    \normalfont B\kern-0.5em{\scshape i\kern-0.25em b}\kern-0.8em\TeX}}}

\begin{document}

\title{Cross-layer Navigation Convolutional Neural Network for Fine-grained Visual Classification}

%
\author{Chenyu Guo$^1$, Jiyang Xie$^1$, Kongming Liang$^1$, Xian Sun$^2$, Zhanyu Ma$^1$}
\affiliation{%
$^1$ Pattern Recognition and Intelligent System Lab., Beijing University of Posts and Telecommunications, Beijing, China\\
$^2$ Aerospace Information Research Institute, Chinese Academy of Sciences, Beijing, China}








\begin{abstract}
  Fine-grained visual classification (FGVC) aims to classify sub-classes of objects in the same super-class (\emph{e.g.}, species of birds, models of cars). For the FGVC tasks, the essential solution is to find discriminative subtle information of the target from local regions. Traditional FGVC models preferred to use the refined features, ~\emph{i.e.}, high-level semantic information for recognition and rarely use low-level information. However, it turns out that low-level information which contains rich detail information also has effect on improving performance. Therefore, in this paper, we propose cross-layer navigation convolutional neural network for feature fusion. First, the feature maps extracted by the backbone network are fed into a convolutional long short-term memory model sequentially from high-level to low-level to perform feature aggregation. Then, attention mechanisms are used after feature fusion to extract spatial and channel information while linking the high-level semantic information and the low-level texture features, which can better locate the discriminative regions for the FGVC. In the experiments, three commonly used FGVC datasets, including CUB-$200$-$2011$, Stanford-Cars, and FGVC-Aircraft datasets, are used for evaluation and we demonstrate the superiority of the proposed method by comparing it with other referred FGVC methods to show that this method achieves superior results. 
  
\end{abstract}





\maketitle

\section{Introduction}

Fine-grained visual classification (FGVC) works to distinguish sub-classes of a common visual super-class (\emph{e.g.}, species of birds, models of cars,~\emph{etc.}). The difference of this task from the general image classification lies in the finer granularity of the classes to which the image belongs. Since different sub-classes differ from each other only in subtle ways, the FGVC tasks face two challenges,~\emph{i.e.}, large intra-class variations and small inter-class difference. Hence, the 
commonly accepted solution is to mine discriminative information as much as possible from local regions for the FGVC tasks.

In this case, in the early work, researchers introduced strong supervisions to mine more discriminative information which requires the manual annotation information of the images, for example, object bounding boxes and part annotations. Such methods first locate distinct regions based on additional annotation information, and then extract features from each of them. However, the fine-grained manual annotations of images are expensive and these methods are rarely used in practice. Subsequently, weakly supervision methods which we used in this paper locate discriminative regions with only class labels. 

One of the representative class of methods used localization-classification subnetworks~\cite{2014The} since every part of the object is important for learning discriminative information in the FGVC. They mainly locate discriminative regions through learning part detectors~\cite{2016Weakly}, attention mechanisms~\cite{2017Look,2017Learning,2018Multi,2017Object,2020Weakly}, or filters/activations \cite{2014The,2020Interpretable}. Among them, the attention mechanism is widely used and the effect is very good. Many models like~\cite{2016SCA,2015Spatial,2018BAM} attempt to incorporate attention mechanisms to improve the performance of CNNs in FGVC tasks. For example, the spatial transformation Network (STN)~\cite{2015Spatial} proposed a learnable module that can be used to locate the most relevant regions in an image to the spatial domain, thus ignoring information from other weakly relevant less important regions. In addition, one way to learn discriminative regions is to integrate multi-level features and guide the features mutually by different levels~\cite{2020Weakly}. 


\begin{figure*}[t]
\centering
\subfigure{
\begin{minipage}[t]{.65\textwidth}
\centering
\includegraphics[width=\linewidth]{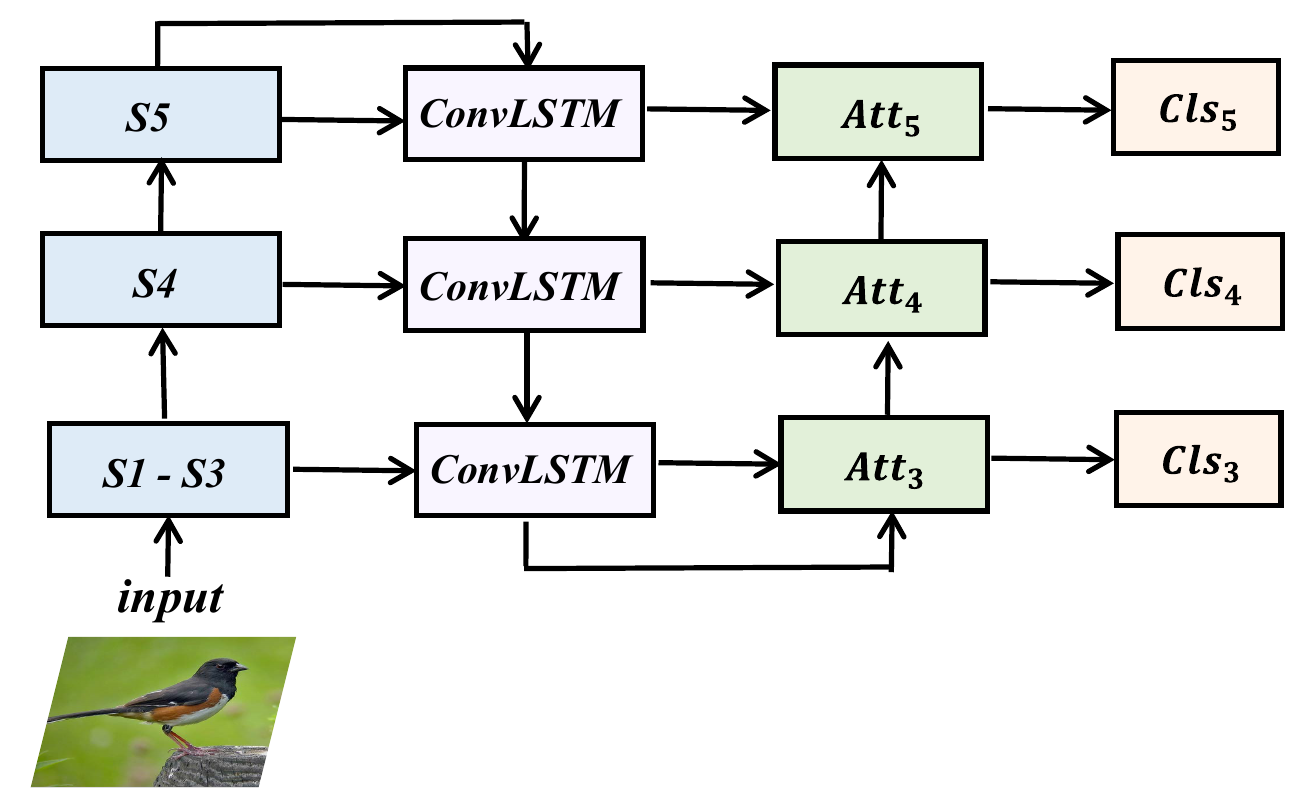} 
\end{minipage}
}
\caption{Overall structure of the proposed CN-CNN. Two-pathway Hierarchy are introduced for cross-layer feature navigation in the FGVC. ``S$\bold1$''-``S$\bold5$'' mean stages in the backbone. ``Att$_i$'' and ``Cls$_i$'' indicate the attention mechanisms in the LH pathway and the classifiers.} \label{fig:structure}
\end{figure*}

Although the above study reports excellent results, the low-level information is sort of overlooked. As the depth of the network increases, the network pays more attention to global high-level abstract semantic information and some low-level detailed information is inescapably lost. In addition, spatio-temporal correlation between layers can refine the process of learning discriminative regions better. Thus, we introduce the ConvLSTM module which can effectively enhance this kind of correlation. In this paper, we propose cross-layer navigation convolutional neural network (CN-CNN), which aims to strengthen the connection between high-level (\emph{i.e.}, semantic information) and low-level (\emph{i.e.}, detailed information) and leverage them for better recognition. First, the feature maps extracted by the backbone network are fed into a convolutional long short-term memory (ConvLSTM) model sequentially from high-level to low-level to perform feature aggregation. Attention mechanisms are used after the feature fusion to extract spatial and channel information while linking high-level semantic information and low-level texture features, which can better locate the discriminative regions for the FGVC.

\begin{figure}[t]
\centering
\subfigure{
\begin{minipage}[t]{.45\textwidth}
\centering
\includegraphics[width=1\linewidth]{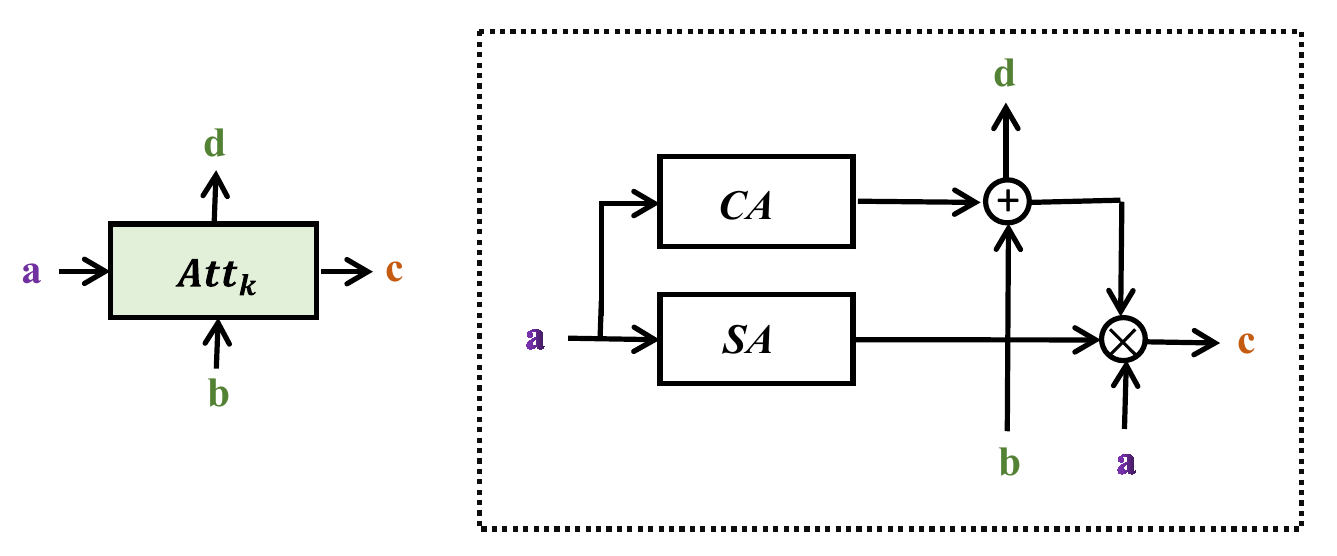}
\end{minipage}
}
\caption{Structure of the attention mechanisms.} \label{fig:att}
\vspace{-6mm}
\end{figure}

The major contributions of this work are summarized as follow: 

\begin{itemize}

\item We propose the CN-CNN, which aims to strengthen the connection between high-level (\emph{i.e.}, semantic information) and low-level (\emph{i.e.}, detailed information) and leverage them for better recognition by ConvLSTM and attention mechanisms. 

\item Experimental results on three common FGVC datasets, including CUB-$200$-$2011$~\cite{2011The}, Stanford-Cars~\cite{20143D}, and FGVC-Aircraft~\cite{2013Fine} datasets show that our proposed method can effectively improve the accuracy in the FGVC.

\end{itemize}

\section{Network Structure}

In this section, we introduce the network architecture of proposed method in detail.
The overall structure of {} is shown in Figure~\ref{fig:structure}.

\subsection{Two-pathway Hierarchy for Cross-layer Navigation}

In this section, we use ResNet$50$ as the backbone for the feature extractor. The feature maps obtained by the third, fourth, and fifth stages are used as the input of the cross-layer navigation, noted as S$3$, S$4$, and S$5$, respectively. Two pathways are proposed, including high-low (HL) pathway, which navigates lower-level features by higher-level ones, and low-high (LH) pathway, which refines higher-level features with lower-level ones.

{\bfseries From high-level to low-level:} The feature maps of different stages are sequentially inputted into ConvLSTM for the guidance from high-level to low-level. The features are aggregated to supervise the underlying texture features using the higher-level semantic features.

{\bfseries From low-level to high-level:} The low-level feature maps then navigate the high-level ones by generating attention maps for them. In the right attention module, there also exists a pathway from low-level to high-level, which is used to deliver the information from the lower layers to the higher ones and link the channels better.

By this two pathway hierarchy structure, the network can well fuse the semantic feature information of the higher-level with the detail information of the lower-level, simultaneously.

\begin{table*}
	\centering
	\fontsize{8}{11}\selectfont    
	\caption{Comparison with state-of-the-art methods on three FGVC datasets. The best results are highlighted in \textbf{bold}.}
	\begin{tabular}{ccccc}
		\toprule
		\toprule
		\multirow{2}{*}{Method}&\multirow{2}{*}{Base}&\multicolumn{3}{c}{Datasets} \cr
		\cmidrule(lr){3-5}& & CUB-$200$-$2011$ & Stanford Cars & FGVC-Aircraft  \cr
		\cmidrule(lr){1-5}
		FT VGGNet (CVPR $18$)~\cite{2018Learning} &VGG$19$ & $77.8$ & $84.9$ & $84.8$   \cr
		FT ResNet (CVPR $18$)~\cite{2018Learning} & ResNet$50$ & $84.1$ & $91.7$ & $88.5$  \cr
		B-CNN (ICCV $15$)~\cite{2015Bilinear}& VGG$16$ & $84.1$ & $91.3$ & $84.1$  \cr
		MA-CNN (ICCV $17$)~\cite{2017Learning} &VGG$19$ & $86.5$ & $92.8$ & $89.9$  \cr
		NTS (ECCV $18$)~\cite{2018}&ResNet$50$ & $87.5$ & $93.9$ & $91.4$  \cr
		Cross-X (ICCV $19$)~\cite{0Cross} &ResNet$50$ & $87.7$ & $94.6$ & $92.6$  \cr
		DCL (CVPR $19$)~\cite{2019Destruction} &ResNet$50$ & $87.8$ & $94.5$ & $93.0$  \cr
		TASN (CVPR $19$)~\cite{9Looking} &ResNet$50$ & $87.9$ & $93.8$ & -  \cr
		CIN (AAAI $20$)~\cite{2020Channel}&ResNet$50$ & $87.5$ & $94.1$ & $92.8$  \cr
		MC-Loss (TIP $20$)~\cite{9005389}&ResNet$50$ & $87.3$ & $93.7$ & $92.6$  \cr
		\hline
	    CN-CNN (ours) &ResNet$50$ & $\boldsymbol{88.9}$ & $\boldsymbol{94.9}$ & $\boldsymbol{94.1}$  \cr
		\bottomrule
		\bottomrule
	\end{tabular}\vspace{0cm}
	\label{tab:sota}
\end{table*}

{\bfseries Classifier:} Finally, the feature maps learned from the two pathway are then fed into three independent classifiers for the FGVC. Each classifier contains a global average pooling (GAP) and two FC layers. The feature maps obtained by the LH pathway are then fed into the three layers to obtain the predictions.


\subsection{ConvLSTM for Navigation from High-level to Low-level}

Through previous work, we found that the results of some models can be effectively improved by introducing low-level features, even for simple aggregations~\cite{2006A}. Beyond that, we inspired by~\cite{2019DIANet} that ConvlSTM is a very powerful module for connecting and integrating multiple layers of information. Therefore, we consider to introduce CONVLSTM module between multi-layer feature map to help feature fusion between different levels.

In our method, the feature maps of different stages obtained from backbone are first fed into corresponding upsampling operations and $1\times1$  convolution layers for obtaining exactly same sizes (\emph{i.e.}, height, width, and channel numbers) of the feature maps. Then they are inputted into ConvLSTM from high-level to low-level. It takes on the role of connecting, guiding and merging different levels of features.

\begin{table}
  \caption{Comparisons of models with/without the two-pathway hierarchy structure. HL: high-low pathway, LH: low-high pathway.}
  \begin{tabular}{ccc}
    \toprule
    Method & Backbone & Accuracy(\%) \\
    \midrule
    \ Baseline & ResNet$50$ & $84.1$ \\
    \ HL & ResNet$50$ & $86.6$ \\
    \ HL+LH & ResNet$50$ & $87.2$ \\
  \bottomrule
\end{tabular}\label{tab:ablation1}
\end{table}

\subsection{Attention Mechanisms for Navigation from Low-level to High-level}

Here, we introduce both spatial and channel attention mechanisms in the navigation from low-level to high-level.

{\bfseries Spatial Attention (SA):} As shown in Figure~\ref{fig:structure}, different feature maps of the distinct levels are refined by attentions. The feature maps are first resized into their original size. Then, they are fed into the SA module, respectively, to obtain the SA masks of different levels. The SA module contains one deconvolution operation with $3\times3$ kernel. The values of the SA masks are normalized into the interval of $[0, 1]$ by Sigmoid function, respectively and they can represent the importance of each region in the specific level.


{\bfseries Channel Attention (CA):} Similarly, with the feature maps as inputs, the CA masks are obtained by a GAP and two FC layers. The channels of the feature maps of different levels are connected to each other and the information of the lower-level is passed from the bottom to the higher-level.

After obtaining both SA and CA masks, we aggregate them together for a pixel-wise masks and multiply the new masks onto the feature maps obtained by the HL pathway.

\begin{table}
  \caption{Comparisons of models with/without the ConvLSTM Module.}
  \begin{tabular}{ccc}
    \toprule
    Method & Backbone & Accuracy(\%) \\
    \midrule
    \ w/o ConvLSTM & ResNet$50$ &  $87.2$ \\
    \ w/ ConvLSTM & ResNet$50$ & $88.9$ \\
  \bottomrule
\end{tabular}\label{tab:ablation2}
\vspace{-6mm}
\end{table}

\section{Experimental Results and Discussion}

We conducted experiments on three common FGVC datasets: Caltech-UCSD Birds (CUB-$200$-$2011$), Stanford-Cars, and FGVC-Aircraft datasets, which are widely used benchmarks for FGVC.

\subsection{Implementation Details}

We conducted all the experiments on GTX $1080$Ti GPU, using Pytorch framework~\cite{paszke2019pytorch}. During training phase, images were resized to $448\times448$ which is a general setting. We used ImageNet-pretrained ResNet$50$~\cite{resnet} as the backbone model. We took stochastic gradient descent (SGD) as the optimizer and used Batch Normalization for regularization. We trained the network for $300$ epochs with initial learning rate as $0.001$ for the backbone and $0.1$ for the other modules. We set the momentum as $0.9$ and the weight decay as $5\times10^{-4}$.

\begin{figure*}[t]
\centering
\subfigure{
\begin{minipage}[t]{.75\textwidth}
\centering
\includegraphics[width=1\linewidth]{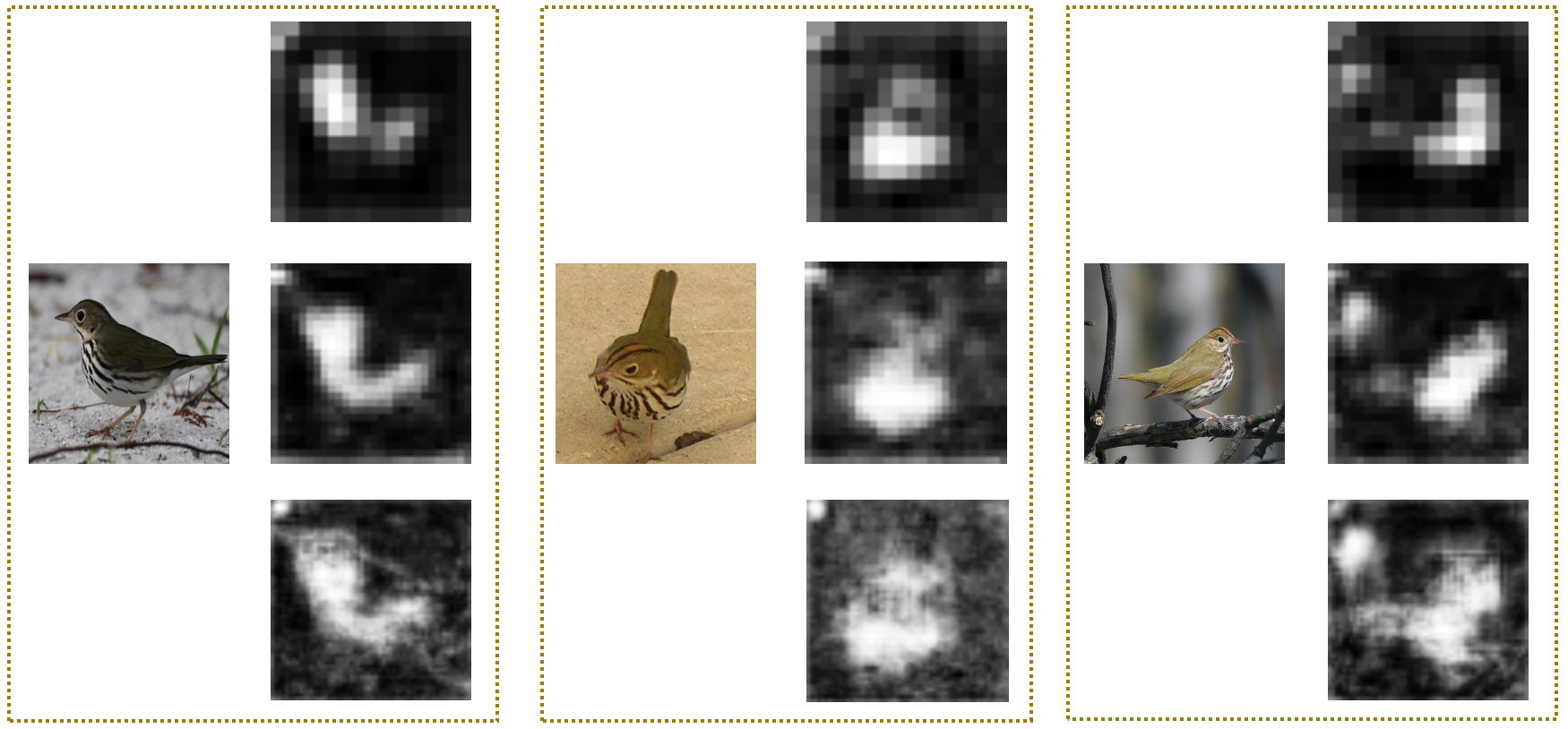}
\end{minipage}
}
\vspace{-4mm}
\caption{Visualization of the output feature maps of the LH pathway. We randomly select three test images from one specific class on the CUB-$\bold200$-$\bold2011$ dataset. The bright white parts in the visualizations is the regions that the network especially pays attention to. In each rectangle, the left color image is the original image. In the right column, from top to bottom, they are high-level to low-level feature maps.}\label{fig:vis}
\end{figure*}

\subsection{Comparisons with State-of-the-Art Methods}

Table~\ref{tab:sota} shows the experiment results on CUB-$200$-$2011$, Stanford Cars, and FGVC-Aircraft datasets, trained with ResNet$50$ as the backbone model. The proposed method I obtains significant improvements on all the three datasets among the referred methods. 

The referred methods listed in Table~\ref{tab:sota} are all weakly supervised methods conducted on the above three datasets. And when using ResNet$50$ as the backbone model to compare with these methods, the proposed method achieves better performance. In other words, it demonstrates the effectiveness of training networks with the cross-layer navigation. Among them, the proposed method produced especially better results on the CUB-$200$-$2011$ dataset that is even more difficult to identify.

\subsection{Ablation Studies}

The ablation studies were conducted on the CUB-$200$-$2011$ dataset with the ResNet$50$ as the backbone model. It aims to find out the impact of each component through comparative experiments.

{\bfseries Impact of the two-pathway hierarchy structure.} To compare the performance of the networks with or without the two-pathway hierarchy structure, we conducted experiments in the baseline, the HL pathway, and the LH pathway. Here, we simply implemented the HL pathway with directly adding the feature maps together, which is same to the operation in feature pyramidal network~\cite{fpn}. In Table~\ref{tab:ablation1}, we can find that there is $2.5\%$ increase in the HL compared with the baseline, which is a significant improvement. Meanwhile, further adding the LH, the accuracy gains another about $0.6\%$ improvement through the enhanced connection between features of different levels. The accuracies in the table clearly demonstrate the benefits of the two-pathway hierarchy structure.

{\bfseries Impact of the ConvLSTM Module.} The difference between the proposed method and baseline (``HL+LH'' in Table~\ref{tab:ablation1}) is that ConvLSTM is introduced. As shown in Table~\ref{tab:ablation2}, it is obvious that the accuracy obtained with the ConvLSTM is much higher than the baseline, with $1.7\%$ performance improvement. This indicates that the ConvLSTM can well enhance the correlation between feature maps. 

\subsection{Visualizations}

Parts of the visualizations of the attention module are shown in Figure~\ref{fig:vis}. We randomly select three test images from one specific class on the CUB-$200$-$2011$ dataset. The bright white parts in the visualizations is the regions that the network especially pays attention to. It can be observed that the low-level information can capture more subtle parts to distinguish the birds. In the higher levels, the network focuses on abstract semantic information more. By integrating low-level information (\emph{e.g.}, colors, edge junctions, and texture patterns), performance can be improved with the enhanced feature representation and the model can accurately locate discriminative regions.

\section{Conclusion}

In this paper, we showed that for the FGVC tasks, it is an effective way to improve classification accuracy by using the cross-layer navigation for feature representation enhancement. We propose cross-layer navigation convolutional neural network (CN-CNN), which aims to strengthen the connection between high-level (\emph{i.e.}, semantic information) and low-level (\emph{i.e.}, detailed information) and leverage them for better recognition. Extensive experiments confirmed that the proposed method improves the accuracy on different FGVC datasets. The effectiveness of the components of the proposed method were also discussed. Visualizations can illustrate the ability of the cross-layer navigation.


\bibliographystyle{ACM-Reference-Format}
\bibliography{ref}


\begin{thebibliography}{27}


\ifx \showCODEN    \undefined \def \showCODEN     #1{\unskip}     \fi
\ifx \showDOI      \undefined \def \showDOI       #1{#1}\fi
\ifx \showISBNx    \undefined \def \showISBNx     #1{\unskip}     \fi
\ifx \showISBNxiii \undefined \def \showISBNxiii  #1{\unskip}     \fi
\ifx \showISSN     \undefined \def \showISSN      #1{\unskip}     \fi
\ifx \showLCCN     \undefined \def \showLCCN      #1{\unskip}     \fi
\ifx \shownote     \undefined \def \shownote      #1{#1}          \fi
\ifx \showarticletitle \undefined \def \showarticletitle #1{#1}   \fi
\ifx \showURL      \undefined \def \showURL       {\relax}        \fi
\providecommand\bibfield[2]{#2}
\providecommand\bibinfo[2]{#2}
\providecommand\natexlab[1]{#1}
\providecommand\showeprint[2][]{arXiv:#2}

\bibitem[\protect\citeauthoryear{Chang, Ding, Xie, Bhunia, Li, Ma, Wu, Guo, and
  Song}{Chang et~al\mbox{.}}{2020}]%
        {9005389}
\bibfield{author}{\bibinfo{person}{Dongliang Chang}, \bibinfo{person}{Yifeng
  Ding}, \bibinfo{person}{Jiyang Xie}, \bibinfo{person}{Ayan~Kumar Bhunia},
  \bibinfo{person}{Xiaoxu Li}, \bibinfo{person}{Zhanyu Ma},
  \bibinfo{person}{Ming Wu}, \bibinfo{person}{Jun Guo}, {and}
  \bibinfo{person}{Yi-Zhe Song}.} \bibinfo{year}{2020}\natexlab{}.
\newblock \showarticletitle{The devil is in the channels: Mutual-channel loss
  for fine-grained image classification}.
\newblock \bibinfo{journal}{\emph{IEEE Transactions on Image Processing}}
  \bibinfo{volume}{29} (\bibinfo{year}{2020}), \bibinfo{pages}{4683--4695}.
\newblock


\bibitem[\protect\citeauthoryear{Chen, Zhang, Xiao, Nie, Shao, Liu, and
  Chua}{Chen et~al\mbox{.}}{2017}]%
        {2016SCA}
\bibfield{author}{\bibinfo{person}{Long Chen}, \bibinfo{person}{Hanwang Zhang},
  \bibinfo{person}{Jun Xiao}, \bibinfo{person}{Liqiang Nie},
  \bibinfo{person}{Jian Shao}, \bibinfo{person}{Wei Liu}, {and}
  \bibinfo{person}{Tat-Seng Chua}.} \bibinfo{year}{2017}\natexlab{}.
\newblock \showarticletitle{Sca-cnn: Spatial and channel-wise attention in
  convolutional networks for image captioning}. In
  \bibinfo{booktitle}{\emph{Proceedings of the IEEE conference on computer
  vision and pattern recognition}}. \bibinfo{pages}{5659--5667}.
\newblock


\bibitem[\protect\citeauthoryear{Chen, Bai, Zhang, and Mei}{Chen
  et~al\mbox{.}}{2019}]%
        {2019Destruction}
\bibfield{author}{\bibinfo{person}{Yue Chen}, \bibinfo{person}{Yalong Bai},
  \bibinfo{person}{Wei Zhang}, {and} \bibinfo{person}{Tao Mei}.}
  \bibinfo{year}{2019}\natexlab{}.
\newblock \showarticletitle{Destruction and construction learning for
  fine-grained image recognition}. In \bibinfo{booktitle}{\emph{Proceedings of
  the IEEE/CVF Conference on Computer Vision and Pattern Recognition}}.
  \bibinfo{pages}{5157--5166}.
\newblock


\bibitem[\protect\citeauthoryear{Ding, Ma, Wen, Xie, Chang, Si, Wu, and
  Ling}{Ding et~al\mbox{.}}{2021}]%
        {2020Weakly}
\bibfield{author}{\bibinfo{person}{Yifeng Ding}, \bibinfo{person}{Zhanyu Ma},
  \bibinfo{person}{Shaoguo Wen}, \bibinfo{person}{Jiyang Xie},
  \bibinfo{person}{Dongliang Chang}, \bibinfo{person}{Zhongwei Si},
  \bibinfo{person}{Ming Wu}, {and} \bibinfo{person}{Haibin Ling}.}
  \bibinfo{year}{2021}\natexlab{}.
\newblock \showarticletitle{AP-CNN: weakly supervised attention pyramid
  convolutional neural network for fine-grained visual classification}.
\newblock \bibinfo{journal}{\emph{IEEE Transactions on Image Processing}}
  \bibinfo{volume}{30} (\bibinfo{year}{2021}), \bibinfo{pages}{2826--2836}.
\newblock


\bibitem[\protect\citeauthoryear{Fu, Zheng, and Mei}{Fu et~al\mbox{.}}{2017}]%
        {2017Look}
\bibfield{author}{\bibinfo{person}{Jianlong Fu}, \bibinfo{person}{Heliang
  Zheng}, {and} \bibinfo{person}{Tao Mei}.} \bibinfo{year}{2017}\natexlab{}.
\newblock \showarticletitle{Look closer to see better: Recurrent attention
  convolutional neural network for fine-grained image recognition}. In
  \bibinfo{booktitle}{\emph{Proceedings of the IEEE conference on computer
  vision and pattern recognition}}. \bibinfo{pages}{4438--4446}.
\newblock


\bibitem[\protect\citeauthoryear{Gao, Han, Wang, Huang, and Scott}{Gao
  et~al\mbox{.}}{2020}]%
        {2020Channel}
\bibfield{author}{\bibinfo{person}{Yu Gao}, \bibinfo{person}{Xintong Han},
  \bibinfo{person}{Xun Wang}, \bibinfo{person}{Weilin Huang}, {and}
  \bibinfo{person}{Matthew Scott}.} \bibinfo{year}{2020}\natexlab{}.
\newblock \showarticletitle{Channel Interaction Networks for Fine-Grained Image
  Categorization}. In \bibinfo{booktitle}{\emph{Proceedings of the AAAI
  Conference on Artificial Intelligence}}, Vol.~\bibinfo{volume}{34}.
  \bibinfo{pages}{10818--10825}.
\newblock


\bibitem[\protect\citeauthoryear{He, Zhang, Ren, and Sun}{He
  et~al\mbox{.}}{2016}]%
        {resnet}
\bibfield{author}{\bibinfo{person}{Kaiming He}, \bibinfo{person}{Xiangyu
  Zhang}, \bibinfo{person}{Shaoqing Ren}, {and} \bibinfo{person}{Jian Sun}.}
  \bibinfo{year}{2016}\natexlab{}.
\newblock \showarticletitle{Deep residual learning for image recognition}. In
  \bibinfo{booktitle}{\emph{Proceedings of the IEEE conference on computer
  vision and pattern recognition}}. \bibinfo{pages}{770--778}.
\newblock


\bibitem[\protect\citeauthoryear{Huang and Li}{Huang and Li}{2020}]%
        {2020Interpretable}
\bibfield{author}{\bibinfo{person}{Zixuan Huang} {and} \bibinfo{person}{Yin
  Li}.} \bibinfo{year}{2020}\natexlab{}.
\newblock \showarticletitle{Interpretable and accurate fine-grained recognition
  via region grouping}. In \bibinfo{booktitle}{\emph{Proceedings of the
  IEEE/CVF Conference on Computer Vision and Pattern Recognition}}.
  \bibinfo{pages}{8662--8672}.
\newblock


\bibitem[\protect\citeauthoryear{Huang, Liang, Liang, and Yang}{Huang
  et~al\mbox{.}}{2020}]%
        {2019DIANet}
\bibfield{author}{\bibinfo{person}{Zhongzhan Huang}, \bibinfo{person}{Senwei
  Liang}, \bibinfo{person}{Mingfu Liang}, {and} \bibinfo{person}{Haizhao
  Yang}.} \bibinfo{year}{2020}\natexlab{}.
\newblock \showarticletitle{Dianet: Dense-and-implicit attention network}. In
  \bibinfo{booktitle}{\emph{Proceedings of the AAAI Conference on Artificial
  Intelligence}}, Vol.~\bibinfo{volume}{34}. \bibinfo{pages}{4206--4214}.
\newblock


\bibitem[\protect\citeauthoryear{Jaderberg, Simonyan, Zisserman, and
  Kavukcuoglu}{Jaderberg et~al\mbox{.}}{2015}]%
        {2015Spatial}
\bibfield{author}{\bibinfo{person}{Max Jaderberg}, \bibinfo{person}{Karen
  Simonyan}, \bibinfo{person}{Andrew Zisserman}, {and} \bibinfo{person}{Koray
  Kavukcuoglu}.} \bibinfo{year}{2015}\natexlab{}.
\newblock \showarticletitle{Spatial transformer networks}.
\newblock \bibinfo{journal}{\emph{arXiv preprint arXiv:1506.02025}}
  (\bibinfo{year}{2015}).
\newblock


\bibitem[\protect\citeauthoryear{Krause, Stark, Deng, and Fei-Fei}{Krause
  et~al\mbox{.}}{2013}]%
        {20143D}
\bibfield{author}{\bibinfo{person}{Jonathan Krause}, \bibinfo{person}{Michael
  Stark}, \bibinfo{person}{Jia Deng}, {and} \bibinfo{person}{Li Fei-Fei}.}
  \bibinfo{year}{2013}\natexlab{}.
\newblock \showarticletitle{3d object representations for fine-grained
  categorization}. In \bibinfo{booktitle}{\emph{Proceedings of the IEEE
  international conference on computer vision workshops}}.
  \bibinfo{pages}{554--561}.
\newblock


\bibitem[\protect\citeauthoryear{Lin, Doll{\'a}r, Girshick, He, Hariharan, and
  Belongie}{Lin et~al\mbox{.}}{2017}]%
        {fpn}
\bibfield{author}{\bibinfo{person}{Tsung-Yi Lin}, \bibinfo{person}{Piotr
  Doll{\'a}r}, \bibinfo{person}{Ross Girshick}, \bibinfo{person}{Kaiming He},
  \bibinfo{person}{Bharath Hariharan}, {and} \bibinfo{person}{Serge Belongie}.}
  \bibinfo{year}{2017}\natexlab{}.
\newblock \showarticletitle{Feature pyramid networks for object detection}. In
  \bibinfo{booktitle}{\emph{Proceedings of the IEEE conference on computer
  vision and pattern recognition}}. \bibinfo{pages}{2117--2125}.
\newblock


\bibitem[\protect\citeauthoryear{Lin, RoyChowdhury, and Maji}{Lin
  et~al\mbox{.}}{2015}]%
        {2015Bilinear}
\bibfield{author}{\bibinfo{person}{Tsung-Yu Lin}, \bibinfo{person}{Aruni
  RoyChowdhury}, {and} \bibinfo{person}{Subhransu Maji}.}
  \bibinfo{year}{2015}\natexlab{}.
\newblock \showarticletitle{Bilinear cnn models for fine-grained visual
  recognition}. In \bibinfo{booktitle}{\emph{Proceedings of the IEEE
  international conference on computer vision}}. \bibinfo{pages}{1449--1457}.
\newblock


\bibitem[\protect\citeauthoryear{Luo, Yang, Mo, Lu, Davis, Li, Yang, and
  Lim}{Luo et~al\mbox{.}}{2019}]%
        {0Cross}
\bibfield{author}{\bibinfo{person}{Wei Luo}, \bibinfo{person}{Xitong Yang},
  \bibinfo{person}{Xianjie Mo}, \bibinfo{person}{Yuheng Lu},
  \bibinfo{person}{Larry~S Davis}, \bibinfo{person}{Jun Li},
  \bibinfo{person}{Jian Yang}, {and} \bibinfo{person}{Ser-Nam Lim}.}
  \bibinfo{year}{2019}\natexlab{}.
\newblock \showarticletitle{Cross-X learning for fine-grained visual
  categorization}. In \bibinfo{booktitle}{\emph{Proceedings of the IEEE/CVF
  International Conference on Computer Vision}}. \bibinfo{pages}{8242--8251}.
\newblock


\bibitem[\protect\citeauthoryear{Maji, Rahtu, Kannala, Blaschko, and
  Vedaldi}{Maji et~al\mbox{.}}{2013}]%
        {2013Fine}
\bibfield{author}{\bibinfo{person}{Subhransu Maji}, \bibinfo{person}{Esa
  Rahtu}, \bibinfo{person}{Juho Kannala}, \bibinfo{person}{Matthew Blaschko},
  {and} \bibinfo{person}{Andrea Vedaldi}.} \bibinfo{year}{2013}\natexlab{}.
\newblock \showarticletitle{Fine-grained visual classification of aircraft}.
\newblock \bibinfo{journal}{\emph{arXiv preprint arXiv:1306.5151}}
  (\bibinfo{year}{2013}).
\newblock


\bibitem[\protect\citeauthoryear{Park, Woo, Lee, and Kweon}{Park
  et~al\mbox{.}}{2018}]%
        {2018BAM}
\bibfield{author}{\bibinfo{person}{Jongchan Park}, \bibinfo{person}{Sanghyun
  Woo}, \bibinfo{person}{Joon-Young Lee}, {and} \bibinfo{person}{In~So Kweon}.}
  \bibinfo{year}{2018}\natexlab{}.
\newblock \showarticletitle{Bam: Bottleneck attention module}.
\newblock \bibinfo{journal}{\emph{arXiv preprint arXiv:1807.06514}}
  (\bibinfo{year}{2018}).
\newblock


\bibitem[\protect\citeauthoryear{Paszke, Gross, Massa, Lerer, Bradbury, Chanan,
  Killeen, Lin, Gimelshein, Antiga, et~al\mbox{.}}{Paszke
  et~al\mbox{.}}{2019}]%
        {paszke2019pytorch}
\bibfield{author}{\bibinfo{person}{Adam Paszke}, \bibinfo{person}{Sam Gross},
  \bibinfo{person}{Francisco Massa}, \bibinfo{person}{Adam Lerer},
  \bibinfo{person}{James Bradbury}, \bibinfo{person}{Gregory Chanan},
  \bibinfo{person}{Trevor Killeen}, \bibinfo{person}{Zeming Lin},
  \bibinfo{person}{Natalia Gimelshein}, \bibinfo{person}{Luca Antiga},
  {et~al\mbox{.}}} \bibinfo{year}{2019}\natexlab{}.
\newblock \showarticletitle{Pytorch: An imperative style, high-performance deep
  learning library}.
\newblock \bibinfo{journal}{\emph{arXiv preprint arXiv:1912.01703}}
  (\bibinfo{year}{2019}).
\newblock


\bibitem[\protect\citeauthoryear{Peng, He, and Zhao}{Peng
  et~al\mbox{.}}{2017}]%
        {2017Object}
\bibfield{author}{\bibinfo{person}{Yuxin Peng}, \bibinfo{person}{Xiangteng He},
  {and} \bibinfo{person}{Junjie Zhao}.} \bibinfo{year}{2017}\natexlab{}.
\newblock \showarticletitle{Object-part attention model for fine-grained image
  classification}.
\newblock \bibinfo{journal}{\emph{IEEE Transactions on Image Processing}}
  \bibinfo{volume}{27}, \bibinfo{number}{3} (\bibinfo{year}{2017}),
  \bibinfo{pages}{1487--1500}.
\newblock


\bibitem[\protect\citeauthoryear{Sun, Yuan, Zhou, and Ding}{Sun
  et~al\mbox{.}}{2018}]%
        {2018Multi}
\bibfield{author}{\bibinfo{person}{Ming Sun}, \bibinfo{person}{Yuchen Yuan},
  \bibinfo{person}{Feng Zhou}, {and} \bibinfo{person}{Errui Ding}.}
  \bibinfo{year}{2018}\natexlab{}.
\newblock \showarticletitle{Multi-attention multi-class constraint for
  fine-grained image recognition}. In \bibinfo{booktitle}{\emph{Proceedings of
  the European Conference on Computer Vision (ECCV)}}.
  \bibinfo{pages}{805--821}.
\newblock


\bibitem[\protect\citeauthoryear{Wah, Branson, Welinder, Perona, and
  Belongie}{Wah et~al\mbox{.}}{2011}]%
        {2011The}
\bibfield{author}{\bibinfo{person}{Catherine Wah}, \bibinfo{person}{Steve
  Branson}, \bibinfo{person}{Peter Welinder}, \bibinfo{person}{Pietro Perona},
  {and} \bibinfo{person}{Serge Belongie}.} \bibinfo{year}{2011}\natexlab{}.
\newblock \showarticletitle{The caltech-ucsd birds-200-2011 dataset}.
\newblock  (\bibinfo{year}{2011}).
\newblock


\bibitem[\protect\citeauthoryear{Wang, Morariu, and Davis}{Wang
  et~al\mbox{.}}{2018}]%
        {2018Learning}
\bibfield{author}{\bibinfo{person}{Yaming Wang}, \bibinfo{person}{Vlad~I
  Morariu}, {and} \bibinfo{person}{Larry~S Davis}.}
  \bibinfo{year}{2018}\natexlab{}.
\newblock \showarticletitle{Learning a discriminative filter bank within a cnn
  for fine-grained recognition}. In \bibinfo{booktitle}{\emph{Proceedings of
  the IEEE conference on computer vision and pattern recognition}}.
  \bibinfo{pages}{4148--4157}.
\newblock


\bibitem[\protect\citeauthoryear{Wolf and Bileschi}{Wolf and Bileschi}{2006}]%
        {2006A}
\bibfield{author}{\bibinfo{person}{Lior Wolf} {and} \bibinfo{person}{Stanley
  Bileschi}.} \bibinfo{year}{2006}\natexlab{}.
\newblock \showarticletitle{A critical view of context}.
\newblock \bibinfo{journal}{\emph{International Journal of Computer Vision}}
  \bibinfo{volume}{69}, \bibinfo{number}{2} (\bibinfo{year}{2006}),
  \bibinfo{pages}{251--261}.
\newblock


\bibitem[\protect\citeauthoryear{Xiao, Xu, Yang, Zhang, Peng, and Zhang}{Xiao
  et~al\mbox{.}}{2015}]%
        {2014The}
\bibfield{author}{\bibinfo{person}{Tianjun Xiao}, \bibinfo{person}{Yichong Xu},
  \bibinfo{person}{Kuiyuan Yang}, \bibinfo{person}{Jiaxing Zhang},
  \bibinfo{person}{Yuxin Peng}, {and} \bibinfo{person}{Zheng Zhang}.}
  \bibinfo{year}{2015}\natexlab{}.
\newblock \showarticletitle{The application of two-level attention models in
  deep convolutional neural network for fine-grained image classification}. In
  \bibinfo{booktitle}{\emph{Proceedings of the IEEE conference on computer
  vision and pattern recognition}}. \bibinfo{pages}{842--850}.
\newblock


\bibitem[\protect\citeauthoryear{Yang, Luo, Wang, Hu, Gao, and Wang}{Yang
  et~al\mbox{.}}{2018}]%
        {2018}
\bibfield{author}{\bibinfo{person}{Ze Yang}, \bibinfo{person}{Tiange Luo},
  \bibinfo{person}{Dong Wang}, \bibinfo{person}{Zhiqiang Hu},
  \bibinfo{person}{Jun Gao}, {and} \bibinfo{person}{Liwei Wang}.}
  \bibinfo{year}{2018}\natexlab{}.
\newblock \showarticletitle{Learning to navigate for fine-grained
  classification}. In \bibinfo{booktitle}{\emph{Proceedings of the European
  Conference on Computer Vision (ECCV)}}. \bibinfo{pages}{420--435}.
\newblock


\bibitem[\protect\citeauthoryear{Zhang, Wei, Wu, Cai, Lu, Nguyen, and Do}{Zhang
  et~al\mbox{.}}{2016}]%
        {2016Weakly}
\bibfield{author}{\bibinfo{person}{Yu Zhang}, \bibinfo{person}{Xiu-Shen Wei},
  \bibinfo{person}{Jianxin Wu}, \bibinfo{person}{Jianfei Cai},
  \bibinfo{person}{Jiangbo Lu}, \bibinfo{person}{Viet-Anh Nguyen}, {and}
  \bibinfo{person}{Minh~N Do}.} \bibinfo{year}{2016}\natexlab{}.
\newblock \showarticletitle{Weakly supervised fine-grained categorization with
  part-based image representation}.
\newblock \bibinfo{journal}{\emph{IEEE Transactions on Image Processing}}
  \bibinfo{volume}{25}, \bibinfo{number}{4} (\bibinfo{year}{2016}),
  \bibinfo{pages}{1713--1725}.
\newblock


\bibitem[\protect\citeauthoryear{Zheng, Fu, Mei, and Luo}{Zheng
  et~al\mbox{.}}{2017}]%
        {2017Learning}
\bibfield{author}{\bibinfo{person}{Heliang Zheng}, \bibinfo{person}{Jianlong
  Fu}, \bibinfo{person}{Tao Mei}, {and} \bibinfo{person}{Jiebo Luo}.}
  \bibinfo{year}{2017}\natexlab{}.
\newblock \showarticletitle{Learning multi-attention convolutional neural
  network for fine-grained image recognition}. In
  \bibinfo{booktitle}{\emph{Proceedings of the IEEE international conference on
  computer vision}}. \bibinfo{pages}{5209--5217}.
\newblock


\bibitem[\protect\citeauthoryear{Zheng, Fu, Zha, and Luo}{Zheng
  et~al\mbox{.}}{2019}]%
        {9Looking}
\bibfield{author}{\bibinfo{person}{Heliang Zheng}, \bibinfo{person}{Jianlong
  Fu}, \bibinfo{person}{Zheng-Jun Zha}, {and} \bibinfo{person}{Jiebo Luo}.}
  \bibinfo{year}{2019}\natexlab{}.
\newblock \showarticletitle{Looking for the devil in the details: Learning
  trilinear attention sampling network for fine-grained image recognition}. In
  \bibinfo{booktitle}{\emph{Proceedings of the IEEE/CVF Conference on Computer
  Vision and Pattern Recognition}}. \bibinfo{pages}{5012--5021}.
\newblock


\end{thebibliography}


\end{document}